\documentclass[review,authoryear]{elsarticle}
\usepackage{amsmath}
\usepackage{algorithm}
\usepackage{algpseudocode}
\usepackage{tabularx}
\usepackage{multirow}
\usepackage{multicol}
\usepackage{booktabs}
\usepackage{amssymb}
\usepackage{color}
\usepackage{url}
\usepackage{pdflscape} %landscape view of pdf pages
\usepackage{afterpage}
\usepackage{capt-of} % or use the larger `caption` package

\usepackage{cite}

\usepackage{nameref}
\usepackage[hidelinks]{hyperref}
\usepackage{cite}
\usepackage{xcolor}
\hypersetup{
    colorlinks,
    linkcolor={red!50!black},
    citecolor={blue!50!black},
    urlcolor={blue!80!black}
}

\usepackage{lineno}

\usepackage{tikz}
\usepackage{graphicx}
\usetikzlibrary{shapes.geometric}
\usetikzlibrary{shapes.arrows}
\makeatletter
\tikzset{my loop/.style =  {to path={
  \pgfextra{}
  [looseness=12,min distance=10mm]
  \tikz@to@curve@path},font=\sffamily\small
  }}  
\makeatletter 
\usepackage{mathtools}

\newcounter{magicrownumbers}

\journal{European Journal of Operational Research}

\begin{document}

\begin{frontmatter}



%%%%%%%%%%%%%%%%%%%%%%%%%%%%%%%%%%%%%%%%%%%%%%%%%%%%%%%%%%%%%%%%%%%%%%%%%%%%%%%%%%%%%%%%%%%
% TITLE
%%%%%%%%%%%%%%%%%%%%%%%%%%%%%%%%%%%%%%%%%%%%%%%%%%%%%%%%%%%%%%%%%%%%%%%%%%%%%%%%%%%%%%%%%%%
\title{Efficient Column Generation for Cell Detection and Segmentation}
\author[upf]{Chong Zhang\corref{mycorrespondingauthor}}
\cortext[mycorrespondingauthor]{Corresponding author}
\ead{chong.zhang@upf.edu}
\author[ae]{Shaofei Wang}
\author[upf,icrea]{Miguel A. Gonzalez-Ballester}
\author[edlab]{Julian Yarkony\corref{mycorrespondingauthor}}
\ead{julian.e.yarkony@gmail.com}

\address[upf]{SimBioSys, DTIC, Universitat Pompeu Fabra, Barcelona, Spain}
\address[ae]{A\&E Technologies, Beijing, China}
\address[icrea]{ICREA, Spain}
\address[edlab]{Experian Data Lab, USA}

%%%%%%%%%%%%%%%%%%%%%%%%%%%%%%%%%%%%%%%%%%%%%%%%%%%%%%%%%%%%%%%%%%%%%%%%%%%%%%%%%%%%%%%%%%%
% ABSTRACT
%%%%%%%%%%%%%%%%%%%%%%%%%%%%%%%%%%%%%%%%%%%%%%%%%%%%%%%%%%%%%%%%%%%%%%%%%%%%%%%%%%%%%%%%%%%
\begin{abstract}

We study the problem of instance segmentation in biological images with crowded and compact cells. We formulate this task as an integer program where variables correspond to cells and constraints enforce that cells do not overlap. To solve this integer program, we propose a column generation formulation where the pricing program is solved via exact optimization of very small scale integer programs. Column generation is tightened using odd set inequalities which fit elegantly into pricing problem optimization. Our column generation approach achieves fast stable  anytime inference for our instance segmentation problems. We demonstrate on three distinct light microscopy datasets, with several hundred cells each, that our proposed algorithm rapidly achieves or exceeds state of the art accuracy.

\end{abstract}


\end{frontmatter}

%\section{}
%\label{}

%%%%%%%%%%%%%%%%%%%%%%%%%%%%%%%%%%%%%%%%%%%%%%%%%%%%%%%%%%%%%%%%%%%%%%%%%%%%%%%%%%%%%%%%%%%
%\textcolor{r}{NOTE: 0) blue colored text are added that are not in miccai15 text (in black colored text), and red colored text are CZ notes; 1) check figures size for current format; 2) avoid using too too many "Note that"!}
%%%%%%%%%%%%%%%%%%%%%%%%%%%%%%%%%%%%%%%%%%%%%%%%%%%%%%%%%%%%%%%%%%%%%%%%%%%%%%%%%%%%%%%%%%%

%%%%%%%%%%%%%%%%%%%%%%%%%%%%%%%%%%%%%%%%%%%%%%%%%%%%%%%%%%%%%%%%%%%%%%%%%%%%%%%%%%%%%%%%%%%
% INTRODUCTION
%%%%%%%%%%%%%%%%%%%%%%%%%%%%%%%%%%%%%%%%%%%%%%%%%%%%%%%%%%%%%%%%%%%%%%%%%%%%%%%%%%%%%%%%%%%
\section{Introduction}
\label{sec:intro}
%Cell detection and segmentation are fundamental tasks for further cell-level quantifications~\citep{Meijering2012}.
Cell detection and instance segmentation are fundamental tasks for the study of bioimages~\citep{Meijering2012} in the era of big data.  Detection corresponds to the problem of identifying individual cells and instance segmentation corresponds to the problem of determining the pixels corresponding to each of the cells.  
 Cells are often in close proximity and/or occlude each other. Traditionally bioimages were manually analyzed, however recent advances in microscope techniques, automation, long-term high-throughput imaging, etc, result in vast amounts of data from biological experiments, making manual analysis, and even many computer aided methods with hand-tuned parameters, infeasible~\citep{Meijering2016,Hilsenbeck2017}. The large diversity of cell lines and microscopy imaging techniques require the development of algorithms for these tasks to perform robustly and well across data sets. 

In this paper we introduce a novel approach for instance segmentation specialized to bioimage analysis designed to rapidly produce high quality results with little human intervention. The technique described in this paper is applicable to images that have crowded and compact cell regions acquired from different modalities and cell shapes, as long as they produce intensity changes at cell boundaries. Such patterns result from several microscopy imaging techniques, such as trans-illumination (e.g. bright field, dark field, phase contrast) and fluorescence (e.g. through membrane or cytoplasmic staining) images. Thus it is specifically suitable for images from which cells are almost transparent.

We formulate instance segmentation as the problem of selecting a set of visually  meaningful cells under the hard constraint that no two cells overlap (share a common pixel).  This problem corresponds to the  classic integer linear programming (ILP) formulation of the set packing problem~\citep{karp} where sets correspond to cells and elements correspond to pixels.  The number of possible cells is very large and can not be easily enumerated.  We employ a column generation approach~\citep{barnprice} for solving the combinatorial problem where the pricing problem is solved via exact optimization of very small scale integer programs (IPs).  Inference is made tractable by relying on the assumption that cells are small and compact.  When needed we tighten the linear programming (LP) relaxation using odd set inequalities~\citep{heismann2014generalization}. The use of odd set inequalities in our context does not destroy the structure of the pricing problem so branch and price \citep{barnprice} is not needed. 

For the purpose of dimensionality reduction we employ the common technique of aggregating pixels into superpixels.  Superpixels \citep{levinshtein2009turbopixels,achanta2012} are the output of a dimensionality reduction technique that groups pixels in a close proximity with similar visual characteristics and is commonly used as a preprocessing for image segmentation\citep{UCM}.  Superpixels provide a gross over-segmentation of the image meaning that they capture many boundaries not in the ground truth but miss very few boundaries that are part of the ground truth.  Hence we apply our set packing formulation on the superpixels meaning that each set corresponds to a subset of the superpixels and each element is a superpixel.  

%\subsection{Contributions}
Our contributions consist of the following:  
\begin{itemize}
\item Novel formulation of cell instance segmentation amenable to the tools and methodology of the operations research community
\item  Structuring our formulation to admit tightening the corresponding LP relaxation outside of branch/price methods
\item Achieve benchmark level results on real microscopy datasets
\end{itemize}
%
%\subsection{Outline}
%

We structure this document as follows.  In Section~\ref{litReview} we consider the related work in the fields of bioimage analysis and operations research. Next in Section~\ref{Cells} we introduce our set packing formulation of instance segmentation and our column generation formulation with its corresponding pricing problem. Next in Sections~\ref{lbUb} and~\ref{lowerplacmain} we consider the production of anytime integral solutions and lower bounds respectively.  In Section~\ref{ssec:cell} we demonstrate the applicability of our approach to real bioimage data sets. Finally we conclude and consider extensions in Section~\ref{conc}.

\section{Related Work}
\label{litReview}

\subsection{Optimization in computer vision and bioimage analysis}
Our work should be considered in the context of methods that are based on cell boundary information and clustering of super-pixels. Relevant methodologies include contour profile pattern~\citep{Kvarnstrom2008,Mayer2013,Dimopoulos2014}, constrained label cost model~\citep{Zhang14a}, correlation clustering~\citep{Zhang14b,Yarkony2015}, structured learning~\citep{Arteta2012,Liu_isbi14,Funke2015a}, and deep learning~\citep{Ronneberger2015}, etc. A comprehensive review can be found in~\citep{Xing2016}. Here we discuss the most relevant work.   

The method of~\citep{Zhang14b} frames instance segmentation as correlation clustering on a planar or nearly planar graph and relies heavily on the planarity of their clustering problem's structure in order to achieve efficient inference. Our work differs from~\citep{Zhang14b} primarily from the perspective of optimization. Notably, our model is not bound by planarity restrictions and instead relies on the assumption that cells are typically small and compact. Therefore, our model is also applicable to 3D segmentation.

In ~\citep{ZhangSchwingICCV2015} the authors use depth to transform instance segmentation into a labeling problem and thus break the difficult symmetries found in instance segmentation. They formulate the optimization as an ILP and solve it using greedy network flow methods~\citep{boykov2001fast}, notably the Quadratic Pseudo-Boolean optimization (QPBO)~\citep{boros2002pseudo,rother2007optimizing}. The approach in~\citep{ZhangSchwingICCV2015} also requires prior knowledge of the number of labels present in the image, which is not realistic for images crowded with hundreds or thousands of cells. In contrast our proposed ILP framework does not require knowledge of the number of objects in the image.

Our inference approach is inspired by \citep{Wang2017} which tackles multi-object tracking using column generation where the corresponding pricing problem is solved using dynamic programming.  This echoes the much earlier operations research work in diverse areas such as vehicle routing \citep{ropke2009branch}, and cutting stock \citep{gilmore1961linear} which use dynamic programming for pricing.  In contrast our pricing problem optimization is solved by many small ILPs, which can be run in parallel which echoes the work in the operations research community of \citep{barahona1998plant}.  
\subsection{Column generation in operations research}
Column generation~\citep{gilmore1961linear,desaulniers2006column,barnprice} is a popular approach for solving ILPs in which compact formulations result in loose LP relaxations where the fractional solution tends to be uninformative.  Here uninformative  means that the fractional solution can not easily be rounded to a low cost integer solution.  Column generation replaces the LP with a new LP over a much larger space of variables which corresponds to a tighter LP relaxation of the ILP~\citep{geoffrion2010lagrangian,armacost2002composite}. The new LP retains the property from the original LP that it has a finite number of constraints.  %Column generation is also be employed in domains where the cardinality of the set of  variables is too large to be considered explicitly in optimization or perhaps even enumerated.  Broadly speaking column generation is used to solve linear programs where the number of primal variables is large and the number of primal constraints is limited.  % most effectively when the number of

To solve the new LP, the dual of the new LP is considered which has a finite number of variables and a huge number of constraints.  Optimization considers only a limited subset of the primal variables, which is initialized as empty, or set heuristically.  Optimization alternates between solving the LP relaxation over the limited subset of the primal variables (called the master problem) and identifying variables that correspond to violated dual constraints (which is called pricing).  Pricing often corresponds to combinatorial optimization which is often an elegant dynamic program which has the powerful feature that  many primal variables (violated dual constraints) are generated at once.  Approaches with dynamic programming based pricing include (but are not limited to) the diverse fields of cutting stock~\citep{gilmore1961linear,gilmore1965multistage}, routing crews~\citep{lavoie1988new,vance1997airline}, and routing vehicles~\citep{ropke2009branch},

Column generation formulations can be tightened using branch-price methods~\citep{barnhart2000using,barnprice,vance1998branch} which is a variant of branch and bound~\citep{land1960automatic} that is structured as to not disrupt the structure of the pricing problem.  %Branch-price is inn is made non-usable by trivial application of branch-bound.  

Column generation has had few applications in computer vision until recently but has included diverse variants of correlation clustering~\citep{HPlanarCC,PlanarCC,yarkony2015next} with applications to image partitioning, multi-object tracking~\citep{Wang2017}, and multi-human pose estimation~\citep{wang2017efficient}.  Column generation in~\citep{HPlanarCC,PlanarCC,yarkony2015next} is notable in that  the pricing problem is solved using the max cut on a planar graph~\citep{maxcutuni,Bar1,Bar2,Bar3} which is known to be polynomial time solvable via a reduction to perfect matching~\citep{fisher2}.% and such a pricing problem is not found in the operations research literature.  
%

%%%%%%%%%%%%%%%%%%%%%%%%%%%%%%%%%%%%%%%%%%%%%%%%%%%%%%%%%%%%%%%%%%%%%%%%%%%%%%%%%%%%%%%%%%%
% METHOD
%%%%%%%%%%%%%%%%%%%%%%%%%%%%%%%%%%%%%%%%%%%%%%%%%%%%%%%%%%%%%%%%%%%%%%%%%%%%%%%%%%%%%%%%%%%
%\begin{methods}

\section{Problem formulation}
\label{Cells}

\begin{table*}[t] \centering
\resizebox{1.\textwidth}{!}{
\begin{tabular}{ l l l l } 
 \hline
 Term & Form &Index & Meaning  \\ [0.5ex] 
 \hline
 $\mathcal{D}$& set & $d$ & set of super-pixels  \\ 
 $\mathcal{Q}$& set  & $q$ & set of cells  \\
 $Q $ &  $\{0,1\}^{|\mathcal{D}|\times |\mathcal{Q}|}$ & $d,q$ & $Q_{dq}=1$ indicates that $d$ in cell $q$\\
  $\Gamma$ & $\mathbb{R}^{|\mathcal{Q}|}$ & $q$ & $\Gamma_q$ is the cost of cell $q$\\
   $\mathcal{C}$ & set & $c$ & set of  triples \\ 
    $\gamma$ & $\{ 0,1\}^{|\mathcal{Q}|}$ & $q$ & $\gamma_q=1$ indicates that cell $q$ is selected.\\%.  later integrality is relaxed\\
    $\theta $ & $ \mathbb{R}^{|\mathcal{D}|}$ & $d$ & $\theta_d$ is the cost of including $d$ in a cell\\
    $\omega $ & $ \mathbb{R}$ & none & $\omega$ is the cost of instancing a cell\\
$\phi $ & $ \mathbb{R}^{|\mathcal{D}|\times |\mathcal{D}|}$ & $d_1,d_2$ & $\phi_{d_1d_2}$ is the cost of including $d_1,d_2$ in the same cell\\
    $V $ & $ \mathbb{R}_{0+}^{|\mathcal{D}|}$ & $d$ & $V_d$ is the volume of super-pixel $d$\\
    $S$ & $ \mathbb{R}_{0+}^{|\mathcal{D}|\times |\mathcal{D}|}$ & $d_1,d_2$ & $S_{d_1d_2}$ is the distance between the centers of super-pixels $d_1,d_2$\\
       $m_V$& $\mathbb{R}_+$ &  none & maximum volume of a cell\\
   $m_R$& $\mathbb{R}_+$ &  none & maximum radius of a cell\\
%$M_v$& \mathbb{R}_{0+}
  $\hat{\mathcal{Q}}$ & set & $q$ & set of  cells generated during column generation\\ 
  $\hat{\mathcal{C}}$ & set & $c$ & set of  triples generated during column generation\\ 
    $\dot{\mathcal{Q}}$ & set & $q$ & set of  cells generated during a given iteration of column generation\\ 
  $\dot{\mathcal{C}}$ & set & $c$ & set of  triples generated during a given iteration of column generation\\ 
  $\lambda$ & $\mathbb{R}_{0+}^{|\mathcal{D}|}$& $d$ & Lagrange multipliers corresponding to super-pixels\\
  $\kappa$ & $\mathbb{R}_{0+}^{|\mathcal{C}|}$& $c$ & Lagrange multipliers corresponding to triples\\
  $x$ & $\{0,1\}^{|\mathcal{D}|}$& $d$ & $x_d=1$ indicates that super-pixel $d$ is included in the  column being generated\\
 [1ex]%&   %. Denotes the part associate with key-point $d$  \\ [1ex] 
 \hline
\end{tabular}
}
\caption{Summary of Notation}
\label{myNotationTable}
\end{table*}
We now discuss our approach in detail. 
Given an image we start with computing a set of super-pixels (generally named super-voxels in 3D), which provides an over-segmentation of cells. These super-pixels are then clustered into ``perceptually meaningful" regions by constructing an optimization problem that either groups the super-pixels into small coherent cells or labels them as background. The solution to this optimization problem corresponds to the globally optimal selection of cells according to our model, which we formulate/solve as an ILP. We consider our model below and summarize the corresponding notation in Table~\ref{myNotationTable}.

\paragraph{\textbf{Definitions}}
\label{def}
Let $\mathcal{D}$ be the set of super-pixels in an image, $\mathcal{Q}$ be the set of all possible cells, and $ G \in \{ 0,1\}^{|\mathcal{D}|\times|\mathcal{Q}|}$ be the super-pixel/cell incidence matrix where $G_{dq}{=}1$ if and only if super-pixel $d$ is part of the cell $q$. We use $S\in \mathbb{R}_{0+}^{|\mathcal{D}|\times|\mathcal{D}|}$ to describe the Euclidean distance between super-pixels; where $S_{d_1d_2}$ indicates the distance between the centers of the super-pixel pair $d_1$ and $d_2$. We use $V \in \mathbb{R}_{+}^{|\mathcal{D}|}$ to describe the area of super-pixels, with $V_d$ being the area of super-pixel $d$. The indicator vector $\gamma  \in \{0,1\}^{|\mathcal{Q}|}$ gives a feasible segmentation solution, where $\gamma_q{=}1$ indicates that cell $q$ is included in the solution and $\gamma_q{=}0$ otherwise.  A collection of cells specified by $\gamma$ is a valid solution if and only if each super-pixel is associated with at most one active cell.  

We use $\Gamma \in \mathbb{R}^{|\mathcal{Q}|}$ to define a cost vector, where $\Gamma_q$ is the cost associated with including cell $q$ in the segmentation. Here we model such a cost with terms $\theta \in \mathbb{R}^{|\mathcal{D}|}$ and $\phi \in \mathbb{R}^{|\mathcal{D}|\times |\mathcal{D}|}$ which are indexed by $d$ and $d_1,d_2$ respectively.  We use $\theta_d$ to denote the cost for including $d$ in a cell and  
$\phi_{d_1d_2}$ to denote the cost for including $d_1$ and $d_2$ in the same cell. We use $\omega \in \mathbb{R}$ to denote the cost of instancing a cell. We now define $\Gamma_q$ in terms of $\theta$, $\phi$ and $\omega$.
\begin{align}
\Gamma_q=\omega+\sum_{d \in \mathcal{D}}\theta_{d}G_{dq}+\sum_{\substack{d_1,d_2\in \mathcal{D}}}\phi_{d_1d_2}G_{d_1q}G_{d_2q}, 
\end{align}

\paragraph{\textbf{Constraints}}
For most biological problems, it is valid to model cells  of a given type as 
having a maximum radius $m_R$ and a maximum area (volume if in 3D) $m_V$. Clearly, $m_V$ and $m_R$ are model defined parameters that vary from one application to another, but they are also often known a-priori. The radius constraint can be written as follows:
\begin{align}
\label{eq:centroid}
\exists[d_*;
G_{d_*q}=1] \quad \mbox{ s.t. }\;  0=\sum_{d_2 \in \mathcal{D}}G_{d_2q}[S_{d_*,d_2}> m_R] \quad \forall q \in \mathcal{Q}.
\end{align}
For any given $q \in \mathcal{Q}$, any argument $d_*\in \mathcal{D}$ satisfying Eq~\ref{eq:centroid} is called an anchor of $q$.   
Similarly, we write the area constraint as follows.
\begin{align}
\label{eq:area}
 m_V\geq \sum_{d \in \mathcal{D}}G_{dq}V_d \quad \forall q \in \mathcal{Q}.
\end{align}

\paragraph{\textbf{ILP formulation}}
\label{cellSegCol}
Given the above variable definitions, we frame instance segmentation as an ILP that minimizes the total cost of the selected cells:
\begin{align}
\label{intobjCell}
\min_{ \substack{\gamma_q \in \{0,1\} \forall q \in \mathcal{Q}\\ \sum_{q \in \mathcal{Q}}G_{dq}\gamma_q \leq 1  \quad \forall d \in \mathcal{D}}}\sum_{q \in \mathcal{Q}}\Gamma_q \gamma_q
=  \min_{ \substack{\gamma \in \{0,1\}^{|\mathcal{Q}|}\\G\gamma \leq 1 }}\Gamma^{\top} \gamma
\end{align}
The effect of our modeling parameters is summarized in Table~\ref{myEffectTable}.
\begin{table*}[t] \centering
\resizebox{0.6\textwidth}{!}{
\begin{tabular}{l l} 
 \hline
 Term & Effect of Positive Offset  \\ [0.5ex] 
 \hline
$\theta $ & Decrease total volume of cells\\
$\phi $ & Fewer pairs of super-pixels in a common cell\\
$\omega $ & Decrease number of cells detected \\
$m_V$ & Increase maximum volume of cells \\
$m_R$ & Increase maximum radius of cells\\
 [1ex]%&   %. Denotes the part associate with key-point $d$  \\ [1ex] 
 \hline
\end{tabular}
}
\caption{Summary of effect of offsetting values by positive offset}
\label{myEffectTable}
\end{table*}

\paragraph{\textbf{Primal and Dual formulation}}
\label{pdform}
The LP relaxation of Eq~\ref{intobjCell} only contains constraints for cells that share a common super-pixel. This generally results in a tight relaxation, although not always. We tighten the relaxation using odd set inequalities~\citep{heismann2014generalization}. Specifically we use odd set inequalities of size three, (called triples), as similarly imposed in~\citep{Wang2017}.  

Triples are defined as follows: for any set of three unique super-pixels (called a triple) 
the number of selected cells of $\mathcal{Q}$ that include two or more of super-pixels in $\{d_1,d_2,d_3\}$ can be no larger than one, i.e. 
\begin{align}
\label{lpcorrect3}
\sum_{q \in \mathcal{Q}}[G_{d_1q}+G_{d_2q}+G_{d_3q} \geq 2]\gamma_q\leq 1.
\end{align}
We denote the set of triples as $\mathcal{C}$ and describe it by a constraint matrix $C {\in} \{0,1\}^{|\mathcal{C}|\times|\mathcal{Q}|}$, where $C_{cq}=1$ if and only if cell $q$ contains two or more members of set $c$. The constraint matrix has a row for each triple:  
$C_{cq}{=}[\sum_{d {\in} c}G_{dq}{\geq} 2], \forall c {\in} \mathcal{C}, q {\in} \mathcal{Q}$.
The primal and dual LP relaxations of instance segmentation with constraints on inequalities corresponding to triples are written below. The dual is expressed using Lagrange multipliers $\lambda \in \mathbb{R}_{0+}^{|\mathcal{D}|}$ and  $\kappa \in \mathbb{R}_{0+}^{|\mathcal{C}|}$.%  We write this below.
\begin{align}
\label{betterRelax}
\min_{\substack{\gamma \geq 0 \\ Q\gamma \leq 1 \\ C\gamma \leq 1 }}\Gamma^{\top}\gamma 
= \max_{\substack{\lambda \geq 0 \\ \kappa \geq 0 \\ \Gamma+Q^{\top}\lambda+C^{\top}\kappa \geq 0}}1^{\top}\lambda +1^{\top}\kappa. 
\end{align}

\subsection{Algorithm}
\label{pdalg}
Since $\mathcal{Q},\mathcal{C}$ are intractably large, we use cutting plane method in the primal and dual to build a sufficient subsets of $\mathcal{Q}$,$\mathcal{C}$.

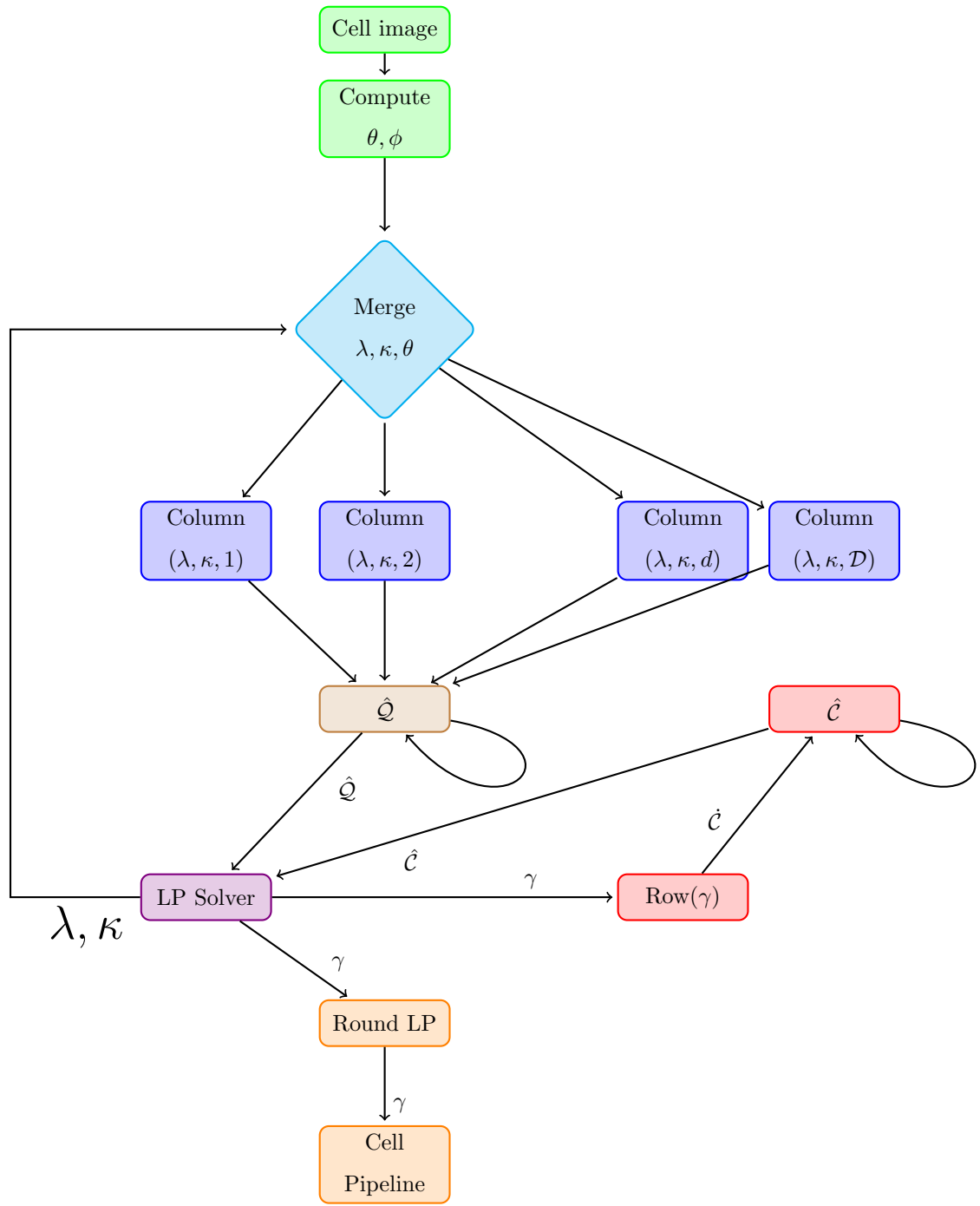
\begin{figure}[t]
%\resizebox{0.5\textwidth}{!}{%
\begin{tikzpicture} [
    auto,
    decision/.style = { diamond, draw=cyan, thick, fill=cyan!20,
                        text width=5em, text badly centered,
                        inner sep=1pt, rounded corners },
    block/.style    = { rectangle, draw=blue, thick, 
                        fill=blue!20, text width=5em, text centered,
                        rounded corners, minimum height=2em },
    Redblock/.style    = { rectangle, draw=red, thick, 
                        fill=red!20, text width=5em, text centered,
                        rounded corners, minimum height=2em },  
    Brownblock/.style    = { rectangle, draw=brown, thick, 
                        fill=brown!20, text width=5em, text centered,
                        rounded corners, minimum height=2em },  
     Greenblock/.style    = { rectangle, draw=green, thick, 
                        fill=green!20, text width=5em, text centered,
                        rounded corners, minimum height=2em }, 
     Orangeblock/.style    = { rectangle, draw=orange, thick, 
                        fill=orange!20, text width=5em, text centered,
                        rounded corners, minimum height=2em },  
     Violetblock/.style    = { rectangle, draw=violet, thick, 
                        fill=violet!20, text width=5em, text centered,
                        rounded corners, minimum height=2em },                                                                                              
    line/.style     = { draw, thick, ->, shorten >=2pt },
  ]
  % Define nodes in a matrix
  \matrix [column sep=3mm, row sep=4mm] {
                    & & \node [Greenblock] (xy) {Cell image};            & \\
                    & & \node [Greenblock] (x) {Compute $\theta,\phi$};            & \\ \\ \\
                    & & \node [decision] (null1) {Merge $\lambda,\kappa ,\theta$};& \\ \\ \\
                    & \node [block] (doa) {Column\\$({\lambda},\kappa,1)$} ; 
                    & \node [block] (doa1) {Column\\$({\lambda},\kappa,2)$};   
                    & \node [block] (doa2) {Column\\$({\lambda},\kappa,d)$};   
                    & \node [block] (doa3) {Column\\$({\lambda},\kappa,\mathcal{D})$};   \\ \\ \\ \\
                    & & \node [Brownblock] (BigBank) {$\hat{\mathcal{Q}}$} edge [in=-50,out=-10,thick,loop] () ;     & & \node [Redblock] (BigBankRow) { $\hat{\mathcal{C}}$} edge [in=-50,out=-10,thick,loop] () ;  &                     \\
    \node(null3){}; & \node [Violetblock] (LPSolver) 
                        {LP Solver}  ; 
                         & & \node [Redblock] (rowFind) {Row$(\gamma)$};   &                         \\ \\ \\
     %                             & \node[text centered](tra){$\mathbf{i}$}; \\
                    & & \node [Orangeblock] (track) {Round LP  }; & \\ \\ \\
                    & & \node [Orangeblock] (FinalSol) {Cell Pipeline}; & \\
  };
  % connect all nodes defined above
  \begin{scope} [every path/.style=line]
    \path (xy)        --    (x);
    \path (x)        --   node [near end] {{ }} (null1); 
    \path (null1)        --   node [near end] { }  (doa);
    \path (null1)        --   node [near end] { }  (doa1);
    \path (null1)        --    node [near end] { } (doa2);
    \path (null1)        --    node [near end] { } (doa3);
    %\path (x)        --    (doa);
   % \path  (LPSolver)   edge[my loop] node[above]  {$e_{1}$} (LPSolver);

    \path (doa)      --    node [near start] { }  (BigBank);%node [near start] {DoAs} (LPSolver);
    \path (doa1)      --  node [near start] { }   (BigBank);%node [near start] {DoAs} (LPSolver);
    \path (doa2)      --  node [near start] { }  (BigBank);%node [near start] {DoAs} (LPSolver);
    \path (doa3)      --  node [near start] { }  (BigBank);%node [near start] {DoAs} (LPSolver);
    \path (BigBank)      --   node [near start] {$\hat{\mathcal{Q}}$}  (LPSolver);%node [near start] {DoAs} (LPSolver);
    \path (BigBankRow)      --  node [near end] {$\hat{\mathcal{C}}$}   (LPSolver);%node [near start] {DoAs} (LPSolver);
    \path (rowFind)      --   node [near start] {$\dot{\mathcal{C}}$}  (BigBankRow);%node [near start] {DoAs} (LPSolver);
    \path (LPSolver) --  node [near end] {{$\gamma$}} (rowFind) ;
  %  \path (tra)      --    (LPSolver);
    \path (LPSolver)   --++  (-3,0) node [near start] {\huge{$\lambda,\kappa $\quad \quad }} |- (null1);
    \path (LPSolver)   --    node [near end] {{$\gamma$}} (track);
    \path (track)   --    node [near end] {$\gamma$} (FinalSol);
  \end{scope}
 \end{tikzpicture}
% } %
\caption{Overview of our system.  We use colors to distinguish between the different parts of the system, which are defined as follows: user input (\textit{green}), sub-problem solution (\textit{blue}), triples (\textit{red}), rounding the output of the LP solver (\textit{orange}).  We use Column$(\lambda,\kappa,d)$ to refer to generating the column where $d$ is an anchor.} 
\label{fig:sys}
\end{figure}

We denote the nascent subsets of $\mathcal{Q},\mathcal{C}$
as $\hat{\mathcal{Q}}, \hat{\mathcal{C}}$ respectively.
In Alg~\ref{cyccell} we write column generation algorithm. We define the cutting plane/column generation in Sections~\ref{rowgen} and~\ref{colgentri} respectively and display optimization in Fig~\ref{fig:sys}.  We use $\dot{\mathcal{Q}}, \dot{\mathcal{C}}$ to refer to the columns and rows generated during a given iteration of our algorithm.  
 \begin{algorithm}[H]
 \caption{Dual Optimization}
\begin{algorithmic} 
\State $\hat{\mathcal{Q}} \leftarrow \{ \}$
\State $\hat{\mathcal{C}} \leftarrow \{ \}$%\quad \hat{\Gamma} \leftarrow \{ \}$  \\
\Repeat
\State $\dot{\mathcal{Q}} \leftarrow \{ \}$
\State $\dot{\mathcal{C}} \leftarrow \{ \}$%\quad \hat{\Gamma} \leftarrow \{ \}$  \\
\State $[\lambda,\kappa,\gamma]\leftarrow$ Solve Primal and Dual of Eq~\ref{betterRelax} over $\hat{\mathcal{Q}},\hat{\mathcal{C}}$%\max_{\substack{\lambda \geq 0\\ 
%\kappa\geq 0 \\
%\Gamma_{\hat{\mathcal{Q}}}+Q_{(:,\hat{\mathcal{Q}})}^{\top}\lambda
%+ C_{(\mathcal{\hat{C}},\mathcal{\hat{Q}})}^{\top}\kappa \geq 0}}
%    -1^{\top}\lambda-1^{\top}\kappa $
%\STATE Recover $\gamma$ from $\lambda,\kappa$
\For {$d_* \in \mathcal{D}$}
\State $q_* \leftarrow \mbox{arg} \min_{\substack{q \in \mathcal{Q} \\ Q_{d_*q}=1\\ 0=\sum_{d_2 \in \mathcal{D}}G_{d_2q}[S_{d_*,d_2}> m_R] }}\Gamma_q+\sum_{d \in \mathcal{D}}Q_{dq}\lambda_{d}+\sum_{c \in \mathcal{C}}\kappa_c[2\leq \sum_{d \in c}Q_{dq}] $
\If{$\Gamma_{q_*}+\sum_{d \in \mathcal{D}}Q_{dq_*}\lambda_{d}+\sum_{c \in \mathcal{C}}\kappa_c[2\leq \sum_{d \in c}Q_{dq}]<0$}
\State $\dot{Q}\leftarrow [\dot{Q} \cup q_*]$
\EndIf
%\STATE $\dot{\mathcal{Q}} \leftarrow \mbox{COLUMN}(\lambda,\kappa) $
\EndFor
%\STATE $ \dot{\mathcal{C}} \leftarrow \mbox{ROW}(\gamma)$
\State $c_*\leftarrow\max_{c \in \mathcal{C}}\sum_{q \in \mathcal{Q}} C_{cq} \gamma_q$
\If{$\sum_{q \in \mathcal{Q}} C_{cq} \gamma_q>1$}
\State $\dot{\mathcal{C}} \leftarrow c_*$
\EndIf
%\IF{ $\Gamma_{p} \neq NULL $}
%
\State  $\hat{\mathcal{Q}}\leftarrow [\hat{\mathcal{Q}},\dot{\mathcal{Q}}]$
\State   $\hat{\mathcal{C}}\leftarrow [\hat{\mathcal{C}},\dot{\mathcal{C}}]$
 \Until{ $\dot{\mathcal{Q}}=[]$  and $\dot{\mathcal{C}}=[]$ }
\end{algorithmic}
\label{cyccell}
  \end{algorithm}

\subsection{Row generation}
\label{rowgen}
Finding the most violated row consists of the following optimization.  
\begin{align}
\max_{c \in \mathcal{C}}\sum_{q \in \mathcal{Q}} C_{cq} \gamma_q
\end{align}
Enumerating $\mathcal{C}$ is unnecessary and we generate its rows as needed by considering
only $c=\{ d_{c_1} d_{c_2} d_{c_3} \}$ such that for each of pair $d_{c_i},d_{c_j}$  there exists an index $q$ such that $\gamma_q>0$ and $Q_{d_iq}=Q_{d_jq}=1$.  Generating rows is done only when no (significantly) violated  columns exist.
%
%At any given time we only consider a subset of the rows and hence only a subset has non -zero value
%
Triples are only added to $\hat{\mathcal{C}}$ if the corresponding constraint is violated.  We can add one or more than one per iteration depending on a schedule chosen.  
\subsection{Generating columns }
\label{colgentri}
Violated constraints in the dual correspond to primal variables (cells) that may improve the primal objective.  To identify such primal variables we compute for each $d_* \in \mathcal{D} $ the most violated dual constraint  corresponding to a cell such that $d_*$ is an anchor of that cell. The corresponding cell is described using indicator vector $x {\in} \{ 0,1\}^{|\mathcal{D}|}$, where the corresponding column is defined as $Q_{dq}\leftarrow x_d, \forall d \in \mathcal{D}$. We write the pricing problem as an IP below.  
\begin{align}
\label{getcolcell}
\min_{\substack{x \in \{ 0,1\}^{|\mathcal{Q}|}}} & \sum_{d \in \mathcal{D}}(\theta_d+\lambda_d)x_d+ 
 \sum_{d_1,d_2 \in \mathcal{D}}\phi_{d_1d_2}x_{d_1}x_{d_2} +\sum_{c \in \mathcal{C}}\kappa_c([2 \leq \sum_{d \in c}x_d]) \nonumber \\
\mathrm{s.t.} \; \; \;  &  x_{d_*}=1  \nonumber \\
& x_d =0 \quad \forall d\in \mathcal{D} \quad \mbox { s.t. }S_{d,d_*}>m_R \nonumber \\
& \sum_{d\in \mathcal{D}}V_dx_d \leq m_V
\end{align}
For our data sets of images crowded with several hundreds of cells, 
the maximum radius of a cell is relatively small and the number of super-pixels within the radius of a given super-pixel is of the order of tens and often around ten. Therefore, solving Eq~\ref{getcolcell} is efficient and can be done in parallel for each $d_* \in \mathcal{D}$. We tackle this by converting Eq~\ref{getcolcell} to an ILP and then solving it with an off-the-shelf ILP solver. %

\begin{figure*}[!t]
\begin{center}
\includegraphics[angle=0, width=0.90\textwidth]{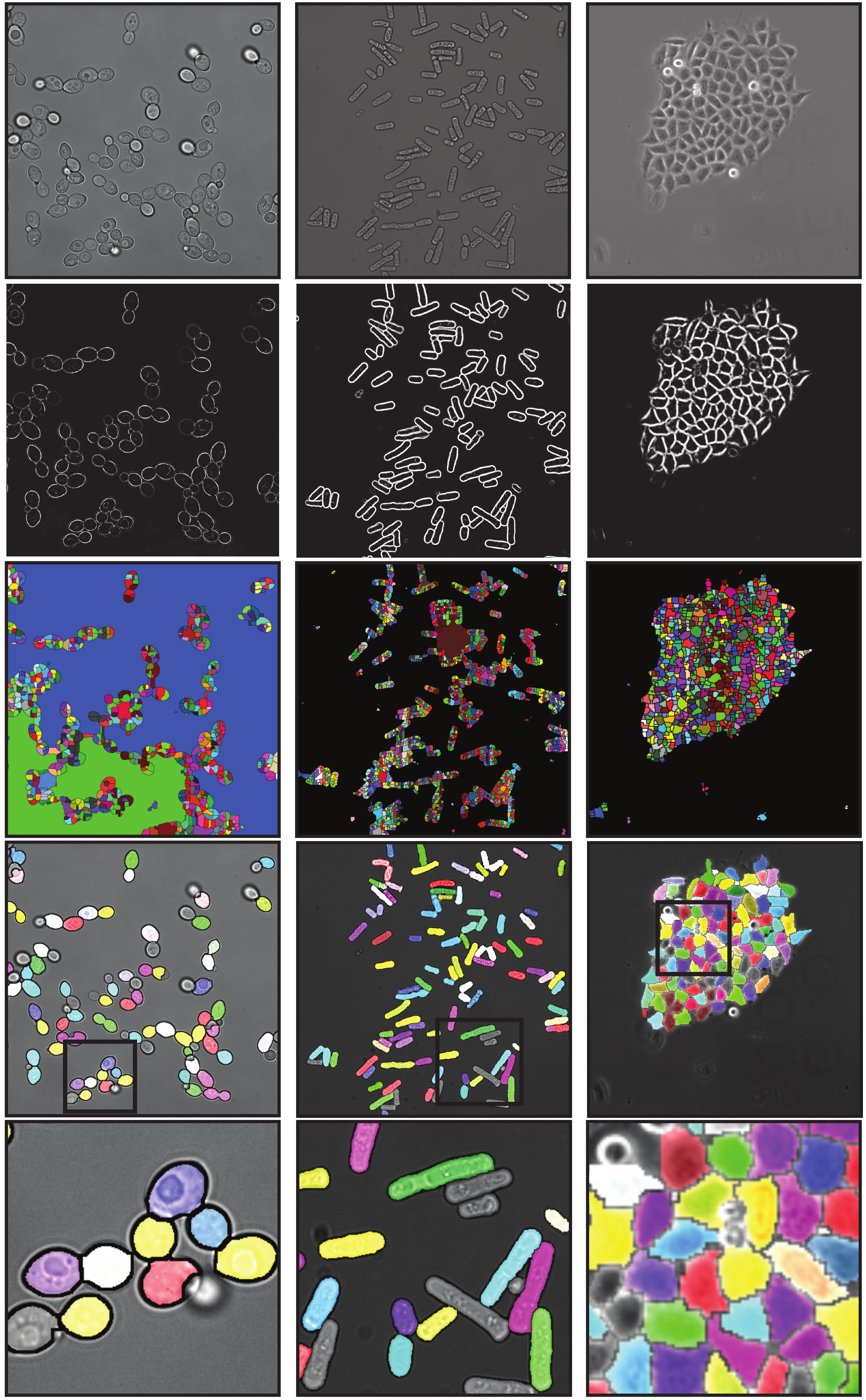}
  %\vspace{5mm}
  \caption[]{Example cell segmentation results of Datasets 1-3 (\emph{left} to \emph{right}). Rows are (\emph{top} to \emph{bottom}): original image, cell of interest boundary classifier prediction image, super-pixels, color map of segmentation, and enlarged views of the inset (\emph{black square})  }
  \label{fig:seg_examples_cells}
  %\vspace{-2mm}
  \end{center}
\end{figure*}
\section{Anytime Integral Solutions}
\label{lbUb}
%In this section we discuss the construction of anytime upper and lower bounds on the optimal integer solution to our problem.   
%
%\paragraph{\textbf{Upper bounds and rounding.}}
We  now consider the anytime  production of  integral solutions in the master problem. While set packing NP hard in general, in practice the LP relaxations are integral at termination and generally integral after each step of optimization. %To ensure this we tightened our bound when needed using odd set inequalities \cite{Wang2017}.  
For cases where the LP is loose, we find that solving the ILP given the primal variables generated takes little additional time beyond solving the LP. However we can use rounding procedures~\citep{Wang2017} when difficult ILPs occur. %For the case of cell segmentation we write the rounding procedure of \cite{Wang2017} below.  
%
%\subsubsection{Rounding Fractional Solutions}
%\label{round}
Specifically, we tackle the rounding of a fractional $\gamma$ with a greedy iterative approach. At each iteration, it selects the cell $q$ with non-binary $\gamma_q$ that minimizes $\Gamma_q\gamma_q$ discounted by the fractional cost of any cells that share a super-pixel with $q$; hence can no longer be added to the segmentation if $q$ is already added.  We write the rounding procedure in Alg~\ref{ubr} using the notation $\mathcal{Q}_{\perp q}$ to indicate
the set of cells in $\mathcal{Q}$ that intersect cell $q$ (excluding $q$
itself).
\begin{algorithm}[!t]
\caption{Upper Bound Rounding}
\begin{algorithmic} 
\While {$\exists q \in \mathcal{Q} \quad \mbox{ s.t. } \gamma_q \notin \{ 0,1 \}$}
\State $q^*\leftarrow \mbox{arg}\min_{\substack{q \in \mathcal{Q} \\ 1>\gamma_q>0}}\Gamma_q\gamma_q -\sum_{\hat{q} \in \mathcal{Q}_{\perp q}} \Gamma_{\hat{q}}\gamma_{\hat{q}} $ \\
\State $\gamma_{\hat{q}} \leftarrow 0 \quad \forall \hat{q} \in \mathcal{Q}_{\perp q^*}$
\State $\gamma_{q^*}\leftarrow 1 $
\EndWhile
\State RETURN $\gamma$
\end{algorithmic}
\label{ubr}
\end{algorithm}
%
%
%%%%%%%%%%%%%%%%%%%%%%%%%%%%%%%%%%%%%%%%%%%%%%%%%%%%%%%%%%%%%%%%%%%%%%%%%%%%%%%%%%%%%%%%%%%%%
\section{Lower bounds}
\label{lowerplacmain}
%
%It is useful in practice to compute a lower bound on the original
%objective during the optimization procedure (i.e., prior to adding all the
%violated columns to the dual).  We rely on the technique of \cite{Wang2017}.   
%
We now consider the production of anytime lower bounds on the optimal integral solution.  
%Computing an anytime lower bound is done using the value of the most violated constraint when generating columns at each iteration. All non-positive values are then summed and added to the value of the LP relaxation to produce a lower bound. This lower bound has value equal to the LP at convergence of column generation since at termination no violated constraints exist. We now derive this result.  
\label{explower}
%When triples are added column generation does not modify the $\ell_s$ with
%relevant modifications to account for $\lambda^{C}$ for all $s \in
%\mathcal{S}$. However it does provide the amount of violation of the most
%violated column. 
We first write the ILP for cell instance segmentation and then introduce Lagrange multipliers.  
 \begin{align}
\min_{\substack{\gamma \in \{0,1\}^{|\mathcal{Q}|} \\ Q\gamma \leq 1 \\ C\gamma \leq 1 }} & \Gamma^{\top}\gamma  = \min_{\substack{\gamma \in \{0,1\}^{|\mathcal{Q}|} \\Q\gamma \leq 1 }}\max_{\substack{\lambda \geq 0 \\ \kappa \geq 0}}\Gamma^{\top}\gamma+(-\lambda^{\top}1+\lambda^{\top}Q\gamma)+(-\kappa^{\top}1+\kappa^{\top}C\gamma)
\label{LBcell0}
 \end{align}
 We now relax the constraint in Eq~\ref{LBcell0} that the dual variables are optimal producing the following lower bound.
 \begin{align}
\nonumber \mbox{Eq }\ref{LBcell0} &\geq  \min_{\substack{\gamma \in \{0,1\}^{|\mathcal{Q}|}\\ Q\gamma \leq 1 }}\Gamma^{\top}\gamma+(-\lambda^{\top}1+\lambda^{\top}Q\gamma)-(\kappa^{\top}1+\kappa^{\top}C\gamma) \\
 &=-\kappa^{\top}1-\lambda^{\top}1+\min_{\substack{\gamma \in \{0,1\}^{|\mathcal{Q}|}\\ Q\gamma \leq 1 }}(\Gamma+Q^{\top}\lambda+C^{\top}\kappa)^{\top}\gamma
% \quad \forall (\lambda \geq 0 , \kappa \geq 0)
 \label{LBcell1}
 \end{align}
Recall that every cell is associated with at least one anchor.  We denote the set of anchors associated with a given cell $q$ as $\mathcal{N}_{q}$.  We use $Q_{:,q},C_{:,q}$ to refer to the column $q$ of the matrices $Q,C$ respectively.  Given any fixed $\gamma\in \{0,1\}^{|\mathcal{Q}|}$ such that $Q\gamma \leq1$ observe the following.  
\begin{align}
(\Gamma+Q^{\top}\lambda+C^{\top}\kappa)^{\top}\gamma \geq \sum_{d \in \mathcal{D}}\min[0,\min_{\substack{q\in \mathcal{Q} \\ d \in \mathcal{N}_q}}\gamma_{q}(\Gamma_q+Q_{:,q}^{\top}\lambda+C_{:,q}^{\top}\kappa)]%\min_{q \in  }\gamma_q(\Gamma_q+Q^{\top}_{:,q}\lambda+C^{\top}\kappa)^{\top}\gamma
\label{boundtool}
\end{align}
%
%Observe Eq \ref{boundtool} is tight if and only if each $q \in \mathcal{Q}$ such  $\gamma_q>0$ that there is exactly one anchor and $\gamma_q=0$ for all $q$ such that $\Gamma_q>0$ and $\gamma_q>0$ implies $\Gamma_q\leq 0$.  %Otherwise it is a lower bound.  
We now use Eq~\ref{boundtool} to produce following lower bound on Eq~\ref{LBcell1}. 
\begin{align}
& \mbox{Eq } \ref{LBcell1}\geq 
& -\kappa^{\top}1-\lambda^{\top}1+\min_{\substack{\gamma {\in} \{0,1\}^{|\mathcal{Q}|}\\ Q\gamma \leq 1 }}\sum_{d \in \mathcal{D}}\min[0,\min_{\substack{q\in \mathcal{Q} \\ d \in \mathcal{N}_q}}\gamma_{q}(\Gamma_q {+} Q_{:,q}^{\top}\lambda {+} C_{:,q}^{\top}\kappa)]
\label{LBcell2}
\end{align}
We now relax the constraint in Eq~\ref{LBcell2} that $Q\gamma \leq 1$ producing the following lower bound.  
\begin{align}
& \mbox{Eq } \ref{LBcell2} 
 \geq {-} \kappa^{\top}1 {-}\lambda^{\top}1 {+} \min_{\substack{\gamma \in \{0,1\}^{|\mathcal{Q}|} }}\sum_{d \in \mathcal{D}}\min[0,\min_{\substack{q\in \mathcal{Q} \\ d \in \mathcal{N}_q}}\gamma_{q}(\Gamma_q {+} Q_{:,q}^{\top}\lambda {+} C_{:,q}^{\top}\kappa)] \nonumber \\
& = -\kappa^{\top}1-\lambda^{\top}1+\sum_{d \in \mathcal{D}}\min[0,\min_{\substack{q\in \mathcal{Q} \\ d \in \mathcal{N}_q}}(\Gamma_q+Q_{:,q}^{\top}\lambda+C_{:,q}^{\top}\kappa)]
%=\kappa^{\top}1-\lambda^{\top}1+\sum_{d \in \mathcal{D}} \min[0,\min_{\substack{q\in \mathcal{Q} \\ d \in \mathcal{N}_q}}(\Gamma_q+Q_{:,q}^{\top}\lambda+C_{:,q}^{\top}\kappa)]]
\end{align}
Observe that the term $\min_{\substack{q\in \mathcal{Q} \\ d \in \mathcal{N}_q}}(\Gamma_q+Q_{:,q}^{\top}\lambda+C_{:,q}^{\top}\kappa)$ is identical to the optimization computed at every stage of column generation. 

%%%%%%%%%%%%%%%%%%%%%%%%%%%%%%%%%%%%%%%%%%%%%%%%%%%%%%%%%%%%%%%%%%%%%%%%%%%%%%%%%%%%%%%%%%%
% RESULTS
%%%%%%%%%%%%%%%%%%%%%%%%%%%%%%%%%%%%%%%%%%%%%%%%%%%%%%%%%%%%%%%%%%%%%%%%%%%%%%%%%%%%%%%%%%%
\section{Results}
\label{ssec:cell}

The technique described in this paper is applicable to images crowded with cells which are mainly discernible by boundary cues. Such images can be acquired from different modalities and cell types. Here we evaluate our algorithm on three datasets. Challenges of these datasets include: densely packed and touching cells, out-of-focus artifacts, variations on shape and size, changing boundaries even on the same cell, as well as other structures showing similar boundaries. 

\subsection{Experiment settings} 
To ensure detecting cell boundaries with varying patterns, a trainable classifier seems to be the right choice. For each dataset, we choose to train a Random Forest (RF) classifier from the open source software, ilastik~\citep{ilastik_0_5}, to discriminate: (1) boundaries of in-focus cells; (2) in-focus cells; (3) out-of-focus cells; and (4) background. The posterior probabilities for class (1) is used as the pairwise potentials. For training, we used $<1\%$ pixels per dataset with generic features e.g. Gaussian, Laplacian, Structured tensor. Subsequent steps use the posterior probabilities to calculate parameters and require no more training. The prediction from the class boundaries of in-focus cells is also used to generate super-pixels. And those for classes (3) and (4) are combined and inverted to create a foreground prediction.  Here foreground corresponds to the superpixels that are part of cells which are background otherwise.  For each super-pixel, the proportion of its foreground part defines the unary potential $\theta$ which we then offset by a constant fixed for each dataset. 
%Specifically, for the studied three distinct sets of microscopy images, they are set to $[-0.4,-0.7]$ and $(-0.5, 0.1, 0.7)$. %Therefore, we expect that these are parameter values that could be generalized. 
A summary about the parameters used in our experiments are shown in Table~\ref{tab:parameters}. %As of $m_V$, $m_R$, they vary from applications, we roughly set them to be (1300, 35), (1500, 70) and (400, 20), respectively.

\afterpage{%
    \clearpage% Flush earlier floats (otherwise order might not be correct)
    \thispagestyle{empty}% empty page style (?)
    \begin{landscape}% Landscape page
\begin{table*}[!t]
\begin{center}
\caption{\label{tab:parameters} Summary of experimental datasets on the number of cells, cell radius, image size, the number of super-pixels and region adjacent graph (RAG) edges.}
\noindent
\begin{tabular*}{1\columnwidth}{@{\extracolsep{\stretch{1}}}*{6}{c}@{}}
  \toprule
Dataset  				& \# cells & avg. cell radius	&  	image size	& \# super-pixels				& \# RAG edges		\\
  \midrule
1~\citep{Zhang14a}  		& 1768		  	& 30		& 	1024$\times$1024			& 1225$\pm$242 				& 3456$\pm$701			\\ 
 \midrule
2~\citep{Peng2013}  	 	& 2340 & 50		& 	1024$\times$1024			& 3727$\pm$2450				& 10530$\pm$7010  	\\ 
\midrule
3~\citep{Arteta2012}  	& 1073		&  20		& 	400$\times$400			& 1081$\pm$364 				& 3035$\pm$1038	\\ 
  \bottomrule                             
\end{tabular*}
\end{center}
\end{table*}

\begin{table*}[!t]
\begin{center}
\caption{\label{tab:comparison}  Evaluation and comparison of detection for Datasets 1-3 (Fig.~\ref{fig:seg_examples_cells}) on precision (P), recall (R), F-score (F), dice coefficient (D) and Jaccard index (J) are reported for the proposed method, as well as those reported in the state-of-the-art methods. Here~\citep{Zhang14b} uses the algorithms planar correlation clustering (PCC) and non-planar correlation clustering (NPCC).}
\noindent
\begin{tabular*}{0.75\columnwidth}{@{\extracolsep{0pt}}*{10}{cccccccccc}@{}}
%\begin{tabular*}{cccccccccccccccc}
\toprule
\multirow{2}{*}{} & \multicolumn{3}{c}{Dataset 1} & \multicolumn{3}{c}{Dataset 2} & \multicolumn{3}{c}{Dataset 3}  \\
\cline{2-4} \cline{5-7} \cline{8-10}  
            & P & R & F             & P & R & F 		        & P & R & F 	  \\
 \midrule
\citep{Arteta2012}            & - & - & -                 & - & - & - 	        & 0.89 & 0.86 & 0.87 	  \\
\citep{Arteta2016}            & - & - & -                & - & - & - 		        & 0.99 & 0.96 & 0.97 	  \\
\citep{Funke2015a}            & 0.93 & 0.89 & 0.91                  & 0.99 & 0.90 & 0.94  		        & 0.95 & 0.98 & 0.97  	  \\
\citep{Hilsenbeck2017}                 & - & - & - 	        & - & - & -          & - & - & 0.97  	  \\
%\citep{Dimopoulos2014}            & - & - & -                 & - & - & - 	        & - & - & - 	  \\
\citep{Ronneberger2015}                 & - & - & - 		        & - & - & -            & - & - & 0.97	  \\
PCC~\citep{Zhang14b}            & 0.95 & 0.86 & 0.90                 & 0.80 & 0.75 & 0.76 		        & 0.92 & 0.92 & 0.92  	  \\
NPCC~\citep{Zhang14b}            & 0.71 & 0.96 & 0.82                & 0.75 & 0.83 & 0.78 		        & 0.85 & \textbf{0.97} & 0.90 	  \\

\textit{Proposed}           & \textbf{0.99}  & \textbf{0.97} & \textbf{0.98}                 & \textbf{1.00} & \textbf{0.94} & \textbf{0.97}  		        & \textbf{1.00} & \textbf{0.97} & \textbf{0.99}	  \\		
\bottomrule                         
\end{tabular*}
\end{center}
\end{table*}

\begin{table*}[!t]
\begin{center}
\caption{\label{tab:comparisonSeg}  Evaluation and comparison of segmentation for Datasets 1-3 (Fig.~\ref{fig:seg_examples_cells}) on dice coefficient (D) and Jaccard index (J) are reported for the proposed method, as well as those reported in the state-of-the-art methods. Here~\citep{Zhang14b} uses the algorithms planar correlation clustering (PCC) and non-planar correlation clustering (NPCC).}
\noindent
\begin{tabular*}{0.6\columnwidth}{@{\extracolsep{0pt}}*{7}{ccccccc}@{}}
%\begin{tabular*}{cccccccccccccccc}
\toprule
\multirow{2}{*}{} & \multicolumn{2}{c}{Dataset 1} & \multicolumn{2}{c}{Dataset 2} & \multicolumn{2}{c}{Dataset 3}  \\
\cline{2-3} \cline{4-5} \cline{6-7}  
            &  D & J                 &  D & J 		        &  D & J 	  \\
 \midrule
%\citep{Arteta2012}            &  - & -                 &  - & - 		        & - & - 	  \\
%\citep{Arteta2016}            &  - & -                 &  - & - 		        & - & - 	  \\
\citep{Funke2015a}            & 0.90 & 0.82                 & 0.90 & 0.83 		        & \textbf{0.84} & 0.73 	  \\
\citep{Hilsenbeck2017}                 & - & - 		        & - & -            & - & \textbf{0.75} 	  \\
\citep{Dimopoulos2014}            & - & 0.87                 & - & - 		        & - & - 	  \\
\citep{Ronneberger2015}                 & - & - 		        & - & -            & - & 0.74 	  \\
PCC~\citep{Zhang14b}            & 0.87 & 0.84                 & \textbf{0.91} & \textbf{0.85} 		        & 0.79 & 0.72 	  \\
NPCC~\citep{Zhang14b}            & 0.86 & 0.89                 & 0.91 & 0.84 		        & 0.80 & 0.70 	  \\

\textit{Proposed}           & \textbf{0.91} & \textbf{0.90}                 & 0.90 & 0.83  		        & 0.82 & 0.71 	  \\		
\bottomrule                         
\end{tabular*}
\end{center}
\end{table*}

    \end{landscape}
    \clearpage% Flush page
}

\subsection{Evaluation} 
A visualization of the results can be seen in Fig~\ref{fig:seg_examples_cells}. Quantitatively, we compare the performance of our algorithm with those reported in the state-of-the-art methods \citep{Arteta2012,Arteta2016,Funke2015a,Hilsenbeck2017,Dimopoulos2014,Ronneberger2015,Zhang14b}, in terms of detection (precision, recall and F-score) and segmentation (Dice coefficient and Jaccard index). For detection, we establish possible matches between found regions and ground truth (GT) regions based on overlap, and find a Hungarian matching using the 
centroid distance as minimizer. Unmatched GT regions are FN, unmatched 
segmentation regions are FP. Jaccard index is computed between the area of true positive (TP) detection regions $R_{tpd}$ and the area of GT region $R_{gt}$: $(R_{tpd}{\cap} R_{gt})/(R_{tpd}{\cup} R_{gt})$. They are summarized in Tables~\ref{tab:comparison} and~\ref{tab:comparisonSeg}. In general, our method achieves or exceed state of the art performance. Additionally, our method requires very little training for the RF classifiers, as opposed to methods like~\citep{Arteta2012,Arteta2016,Funke2015a}, which require fully labeled data for training. This is an advantage of relieving human annotations when several hundreds of cells need to be labeled per image. Also, our method can handle very well large variations of cell shape/size even in the same image, as shown in Fig~\ref{fig:seg_examples_cells} for Dataset 2.

\subsection{Timing and bounds} 
\begin{figure}
\begin{center}
\includegraphics[clip,trim=.5cm 7cm 1.5cm 7cm,width=0.70\textwidth]{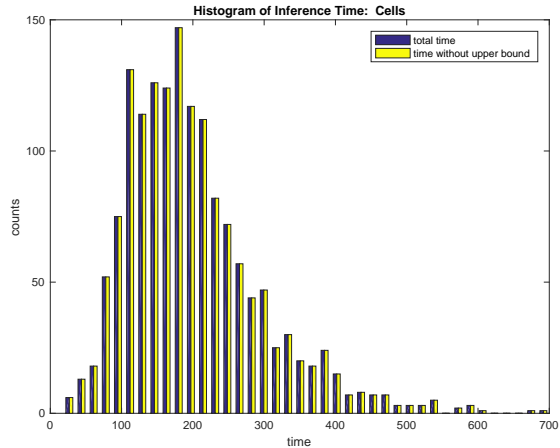}
\end{center}
\caption{Histogram of inference time for Dataset 1.} 
\label{histplotcell}
\end{figure}

We now consider the performance of our approach with regard to the gap between the upper and lower bounds produced by our algorithm.  We normalize these gaps by dividing by the absolute value of the lower bound.  For our three data set the proportion of problem instances that achieve normalized gaps under 0.1 are  99.28 \%, 80 \% and 100 \%, on Datasets 1,2,3 respectively. The peak histogram of inference time are around 150, 500 and 100 seconds without parallelization. As an example, the histogram of inference time for Dataset 1 is shown in Fig~\ref{histplotcell}. Our approach is approximately an order of magnitude faster than that of~\citep{Zhang14b}. 

%%%%%%%%%%%%%%%%%%%%%%%%%%%%%%%%%%%%%%%%%%%%%%%%%%%%%%%%%%%%%%%%%%%%%%%%%%%%%%%%%%%%%%%%%%%
% CONCLUSION
%%%%%%%%%%%%%%%%%%%%%%%%%%%%%%%%%%%%%%%%%%%%%%%%%%%%%%%%%%%%%%%%%%%%%%%%%%%%%%%%%%%%%%%%%%%
\section{Conclusion}
\label{conc}
In this article we introduce a novel column generation strategy that efficiently optimizes an ILP formulation of instance segmentation through clustering super-pixels. We use our approach to detect and segment crowded clusters of cells in distinct microscopy image datasets and achieves state of the art or near state of the art performance.  

We now consider some extensions of our approach. The use of odd set inequalities may prove useful for traditional set cover formulations of vehicle routing problems.  In this context for triples the corresponding inequality is defined as follows: \textit{For any set of three unique depots the number of routes that pass through one or two or those depots plus two times the number of routes that pass through all three depots is no less than two}.  Dynamic programming formulations for pricing can be adapted to include the corresponding Lagrange multipliers~\citep{irnich2005shortest} ( in either the elementary or non-elementary~\citep{kallehauge2005vehicle} setting).  The approach in~\citep{Wang2017} can also be adapted which employs dynamic programming in a branch and bound context in the pricing problem (never the master problem).  

Another extension considers multiple types of cells with a unique model for each cell type and its own pricing problem.  Such types can include rotations, scalings, or other transformations of a common model which may be useful for cells that highly non-circular in shape.  %  Similarly in order to consider diverse ranges of shapes one can try multiple pricing problem each associated with a particular scale and orientation (for various shaped cells). 

In future work one should apply dual feasible inequalities~\citep{ben2006dual,HPlanarCC}. In this case one would create separate variable for each pair of  cell, feasible anchor for that cell (where the anchor is called the the main anchor). Then the ILP would be framed as selecting a set of cells such that (1) no super-pixel is included more than once and (2) no main anchor is included in more than one cell.  However the Lagrange multipliers for (1) can be  bounded from above by the increase in cost corresponding to removing the super-pixel $d$ from a cell. For a super-pixel $d$ one trivial such bound is minus one times the  sum of the non-positive cost terms involving $d$.%(including $\omega$ if $\omega$ is non-positive).  

\section*{Acknowledgment}

This work was partly supported by the Spanish Ministry of Economy and Competitiveness under the Maria de Maeztu Units of Excellence Programme (MDM-2015-0502).\vspace*{-12pt}

\bibliographystyle{model2-names}
\bibliography{cellsegmentation}

%% not want to use model2-names.bst.



%% \bibitem must have one of the following forms:
%%   \bibitem[Jones et al.(1990)]{key}...
%%   \bibitem[Jones et al.(1990)Jones, Baker, and Williams]{key}...
%%   \bibitem[Jones et al., 1990]{key}...
%%   \bibitem[\protect\citeauthoryear{Jones, Baker, and Williams}{Jones
%%       et al.}{1990}]{key}...
%%   \bibitem[\protect\citeauthoryear{Jones et al.}{1990}]{key}...
%%   \bibitem[\protect\astroncite{Jones et al.}{1990}]{key}...
%%   \bibitem[\protect\citename{Jones et al., }1990]{key}...
%%   \harvarditem[Jones et al.]{Jones, Baker, and Williams}{1990}{key}...
%%


% \end{thebibliography}

\end{document}